\documentclass[letterpaper]{article} 
\usepackage{aaai2026}    
\usepackage{times}                   
\usepackage{helvet}                  
\usepackage{courier}                 
\usepackage[hyphens]{url}            
\usepackage{graphicx}                
\usepackage{natbib}                  
\usepackage{caption}                 
\usepackage{algorithm}
\usepackage{algorithmic}
\usepackage{newfloat}
\usepackage{listings}
\DeclareCaptionStyle{ruled}{labelfont=normalfont,labels ep=colon,strut=off}
\floatstyle{ruled}
\newfloat{listing}{tb}{lst}
\floatname{listing}{Listing}

\usepackage{amsmath}
\usepackage{amsfonts}
\usepackage{booktabs}
\usepackage{nicefrac}
\usepackage{array}
\usepackage{tabularx}
\usepackage{makecell}
\usepackage{lipsum} 

\usepackage{placeins}     
\usepackage{subcaption}    
\graphicspath{{pictures/}} 

\graphicspath{{media/}}

\newif\ifanonymous
\anonymousfalse 

\title{Benchmarking IoT Time-Series AD with Event-Level Augmentations}

\ifanonymous
\author{Anonymous AAAI-26 Submission}
\affiliations{Paper under double-blind review}
\fi

\ifanonymous\else
\author{
    Dmitry Zhevnenko\textsuperscript{\rm 1},
    Aleksandr Kovalenko\textsuperscript{\rm 1},
    Fedor Meshchaninov\textsuperscript{\rm 1},
    Anton Kozhukhov\textsuperscript{\rm 2},
    Vladislav Travnikov\textsuperscript{\rm 2},
    Makar Ippolitov\textsuperscript{\rm 2},
    Kirill Yashunin\textsuperscript{\rm 2},
    Iurii Katser\textsuperscript{\rm 2},
    Ilya Makarov\textsuperscript{\rm 1, 3, 4}
}
\affiliations{
    \textsuperscript{\rm 1}AXXX, Moscow, Russia\quad
    \textsuperscript{\rm 2}NSU, Novosibirsk, Russia \quad
    \textsuperscript{\rm 3}Research Center of the Artificial Intelligence Institute, Innopolis University, Innopolis, Russia \quad
    \textsuperscript{\rm 4}Trusted AI Center, RAS, Moscow, Russia \quad
}
\fi

\begin{document}
\maketitle

\begin{abstract}
Anomaly detection (AD) for safety-critical IoT time series should be judged at the event level—by reliability and earliness under realistic perturbations. Yet many studies still emphasize point-level results on base form of curated public datasets, which limits their value for model selection in practice. We introduce an evaluation protocol that adds unified event-level augmentation simulating real-world problems by (i) adds a calibrated sensor dropout stress, linear/log drift, additive noise, and window shifts, and (ii) performs sensor-level probing via mask-as-missing zeroing with per-channel influence estimation to support root-cause analysis.

We evaluate 14 representative models on five public anomaly datasets (SWaT, WADI, SMD, SKAB, TEP) and two industrial datasets (steam turbine, nuclear turbogenerator) under data unified splits and event aggregation. There is no universal winner: graph-structured models transfer best under dropout and long events, for example on SWaT with additive noise a graph autoencoder falls from 0.804 to 0.677 (-16\%) while a graph-attention variant goes from 0.759 to 0.680 (-10\%) and a hybrid graph attention stays nearly flat at 0.762 to 0.756 (-0.8\%); density or flow models work well on clean, stationary plants yet can be fragile to monotone drift, as log drift collapses a flow-based detector on SKAB and nuclear power plant (NPP) dataset while on SWaT the drop is mild at 0.795 to 0.783 (-1.5\%); spectral CNNs lead when periodicity is strong; reconstruction autoencoders become competitive after basic sensor vetting; predictive or hybrid dynamics help when faults break temporal dependencies but remain window-sensitive. The same protocol also informs architecture choices, for example on SWaT under log perturbations replacing normalizing flows with a Gaussian density estimator reduces F1 to about 0.57 at high stress versus about 0.75 originally, and fixing the learned DAG yields a small clean-set gain of about 0.5 to 1.0 points but increases drift sensitivity by roughly eight times.

We release all dataset experiments data and code in Supplementary materials to support reproducible, robustness-oriented evaluation of spatio-temporal AD in IoT settings.
\end{abstract}

\section{Introduction}
Reliable anomaly detection (AD) in complex, safety-critical systems (energy, aerospace, large machinery) must withstand sensor faults, noise, and regime shifts \cite{mallioris2024predictive}. In practice, operators act on \emph{events} rather than isolated points: the core questions are whether an abnormal episode is detected at all, how early it is flagged (when a latency window is defined), and how robust the detector remains when conditions change. Crucially, when a sensor fails there is no opportunity to calibrate at test time; the model must operate under \emph{zero test-time calibration}. Yet prevailing benchmarks optimize pointwise averages on curated data, which obscures event-level behavior, overstates nominal accuracy, and misguides model selection for deployment.

\paragraph{Event view and definitions.}
We formalize anomalies and misleading sensor perturbations as contiguous spatiotemporal \emph{events} in cyber–physical systems (CPS) and adopt an \emph{event-level} evaluation protocol with fixed decision thresholds selected on validation.

\paragraph{Protocol overview.}
We propose a deployment-first evaluation protocol with three components: (i) \emph{base benchmarking} on clean data under data unified splits and event aggregation; (ii) an \emph{offline-calibrated} stress suite that emulates realistic, uncalibrated-at-test-time perturbations — including \emph{sensor dropout}, linear/log drift, additive noise, and window/phase shifts — where severity levels are normalized to per-dataset validation statistics (e.g., relative to nominal variance) and frozen before testing; and (iii) \emph{sensor-level probing} via mask-as-missing zeroing to estimate per-channel influence for root-cause analysis and sensor vetting.

\paragraph{Scope and evidence.}
We evaluate 14 representative AD models across five public CPS datasets (SWaT, WADI, SMD, SKAB, TEP) and two proprietary industrial telemetry sets (steam turbine and nuclear turbogenerator) under identical splits, event aggregation, and stressors. The protocol exposes regime-dependent behavior that \emph{changes} benchmark-driven model choice, with no universal winner. For example, on SWaT with additive noise a graph autoencoder drops from 0.804 to 0.677 (-16\%), a graph-attention variant goes from 0.759 to 0.680 (-10\%), while a hybrid graph–attention remains nearly flat at 0.762 to 0.756 (-0.8\%). Under log drift, flow-based density models can collapse on SKAB and NPP, whereas on SWaT the effect is mild at 0.795 to 0.783 (-1.5\%). Simple sensor zeroing in an industrial run raises F1 from 0.38 to 0.58 (+54\%). These effects persist without any test-time calibration, underscoring the need to match inductive biases to plant stress profiles before deployment.

\paragraph{Positioning.}
Rather than another leaderboard, we provide \emph{design rules} and a reproducible protocol: choose graph-structured models when dropout or long events dominate; prefer density/flow on stable, stationary plants while monitoring drift sensitivity; use spectral CNNs for pronounced periodicity; make reconstruction autoencoders competitive via minimal sensor vetting; and deploy predictive/hybrid dynamics when anomalies break temporal dependencies, acknowledging window sensitivity. We release stress scripts and configurations sufficient to reproduce all public-data results; industrial findings are reported as anonymized aggregates.

\vspace{1mm}
\noindent\textbf{Contributions.}
\begin{itemize}
\item A deployment-oriented, event-level protocol with an offline-calibrated stress suite that enforces zero test-time calibration and includes sensor-level probing for root-cause analysis.
\item A unified study of 14 models on seven CPS datasets under identical splits, event aggregation, and common stressors, revealing regime-dependent reversals in model ranking.
\item Practical design rules mapping stress profiles to model families to support robust, calibration-free operation at inference time.
\item Reproducible artifacts: stress scripts, configs, and seeds for public datasets; industrial results reported as anonymized aggregates.
\end{itemize}

\section{Related work}

\paragraph{Benchmarks and protocols.}
The Numenta Anomaly Benchmark (NAB) introduced latency-aware scoring via anomaly windows and time-weighted rewards, but it targets predominantly \emph{univariate} streams and does not prescribe multivariate cyber–physical systems (CPS) event aggregation or sensor diagnostics \cite{Lavin2015NAB}. The Anomaly Detection Benchmark (ADBench) broadens algorithm and dataset coverage; by default, however, evaluation is \emph{point-level} rather than CPS \emph{event}-centric and does not fix latency or calibration policies \cite{Han2022ADBench}. The Time-Series Benchmark for Unsupervised Anomaly Detection (TSB-UAD) is an end-to-end suite for \emph{univariate} time-series anomaly detection (TSAD) \cite{Paparrizos2022TSBUAD}, and TimeEval is a toolkit to run many detectors, but neither defines a CPS-focused, event-level, \emph{stress-calibrated} protocol \cite{Wenig2022TimeEval}. Prior work has also shown that \emph{point-adjust} counting any hit within an anomaly range as a true positive can inflate scores by masking timing errors, and at the same time, does not use this window to build new event-level testing criteria; range/event-based metrics were proposed to reflect episode-level detection and latency \cite{Wu2023FlawedBenchmarks,Tatbul2018RangePR}. In contrast, we adopt \emph{offline-calibrated stressors} (severity normalized to validation statistics and fixed before testing) and \emph{zero test-time calibration} (no parameter tuning after fault onset).

\paragraph{CPS datasets and common practices.}
Secure Water Treatment (SWaT ), Water Distribution (WADI), and the Tennessee Eastman Process (TEP) are de-facto CPS benchmarks \cite{Goh2016SWAT,Ahmed2017WADI,DownsVogel1993TEP}. Published results frequently combine per-point F1 with post-hoc point-adjust or ad-hoc episode merging; stress tests (sensor dropout, drift, noise) appear, but are rarely specified as \emph{offline-calibrated} or evaluated under \emph{zero test-time calibration}. Surveys summarize deep TSAD families and open issues, yet typically stop short of an operational, deployment-first protocol for multivariate time series in CPS (MVTS-CPS) \cite{zhou2022timeseries}.

\paragraph{Model families for benchmarking.}
We group prior work into five cohorts with complementary \emph{inductive biases} aligned to our stressors (sensor dropout, drift, additive noise, shifts). (i) \emph{Reconstruction} models score deviations by rebuild error: low-rank/sparse decompositions (Robust PCA) \cite{robustpca}, autoencoders for time series \cite{sakurada2014autoencoder}, adversarial two-decoder designs (USAD) \cite{audibert2020usad}, variational LSTM encoders–decoders (LSTM-VAE) \cite{lstm_vae}, and representation/contrastive variants (e.g., DCdetector) \cite{zhou2022contrastive,dcdetector}. They provide strong, inexpensive baselines with interpretable error maps; they tolerate moderate noise but are sensitive to corrupted (\emph{toxic}) channels, motivating sensor vetting. (ii) \emph{Predictive/hybrid} methods model dynamics or combine local/global dependencies: ARIMA/SARIMA \cite{arima,sarima}, convolutional predictors for TSAD (DeepAnt) \cite{deepant}, self-attention predictors \cite{kim2023time}, LSTM with nonparametric dynamic thresholds (LSTM-NDT) \cite{lstm_ndt}, the Anomaly Transformer with association-discrepancy loss \cite{anomaly_transformer}, and MTAD-GAT with parallel temporal/variable attention \cite{zhao2020multivariate}. They often reduce latency when faults break temporal structure but are window/lag-sensitive and can degrade under heavy dropout without explicit variable attention. (iii) \emph{Spectral/seasonal CNNs} (TimesNet) map 1D signals to frequency-aware 2D variations guided by FFT, excelling under stable seasonality while being brittle to sensor shifts \cite{timesnet,fft_anom}. (iv) \emph{Graph-structured} models make inter-sensor topology explicit: GDN learns variable graphs with attention forecasting \cite{deng2021graph}, GBAD adds learnable adjacency to a GCN encoder \cite{gbad}, and GTA couples Gumbel-Softmax topology learning with Transformer-based temporal modeling \cite{chen2021learning}. Such priors help under block dropout and long events, with complexity growing with sensor count and possible misses on narrow single-channel spikes. (v) \emph{Density/flow} likelihoods estimate plausibility directly: THOC builds hierarchical one-class temporal representations \cite{thoc}, while GANF uses DAG factorization with normalizing flows for nonlinear densities \cite{ganf}; these are compelling on clean, stationary plants but can be fragile under misspecified monotone transforms and certain drifts. This coverage links stress-profile sensitivities (dropout/drift/noise/periodicity) to family-level biases, enabling rigorous, event-level comparisons rather than headline point-metrics.

\section{Datasets}

We selected five public datasets from real industrial processes to cover a range of applications. We included two proprietary datasets from critical energy industries: one for steam turbine monitoring in nuclear power plants and one for turbo-generator monitoring. For each dataset, we reported its short description here, and extended version with data sample provided is presented in Supplementary materials.

Unless explicitly stated otherwise in the dataset-specific notes below, the following defaults apply. Anomaly detection is treated as a \emph{window–level} classification task, with labels taken directly from the original dataset authors. For datasets providing point or interval labels, a sliding window is assigned a positive label if it overlaps a labeled anomalous interval at least at one timestamp; otherwise negative. Transitions are not marked as anomalies, and every attack or event is considered anomalous. No latency window is applied, and the train/test split follows the authors' division. The window length depends on the specific model used for training, and the complete original dataset is utilized in all cases, \emph{except for documented dataset-specific adjustments}.

\subsection{Description
\footnote{We thank iTrust, Centre for Research in Cyber Security, Singapore University of Technology and Design, for providing access to the SWaT  and WADI datasets.}
}

\paragraph{SWaT} The SWaT   dataset \cite{mathur2016swat}, version A2, represents data collected from sensors of the testbed of the water treatment plant of the same name. Data was captured from sensors and actuators (51 features overall)  at a frequency of 1 Hz. During data collection, the first 7 days normal operation data were recorded (train dataset), and the last 4 days comprise a sequence of 41 attacks on physical components and software (test dataset), with the anomaly rate 12.14\%.  

\paragraph{WADI} \emph{(Overrides defaults: non-informative/empty columns removed; see Preprocessing.)} The WADI (Water Distribution) dataset \cite{goh2017dataset}, version A2, is a set of data collected from the sensors of a miniature water distribution system connected to the SWaT  bench. Data for 123 sensor or actuator readings were collected over 16 consecutive days at 1 Hz. The logs contain long gap ($\approx$ 2 days) in the middle of this period. The authors performed 15 attacks on the system in the last 2 days to create the test part of the dataset.  The anomaly rate equals to 5.77\% , the mean attack length is about 11 minutes, and a combined duration of attacks constitute 166 minutes 20 seconds.

\paragraph{SMD} Server Machine Dataset \cite{su2019robust} is an anonymized dataset of the three server groups functioning over 5 weeks. The dataset consists of 28 multivariate time-series related to different servers. The first temporal half of the dataset (approximately 708400 timestamps) corresponds to regular operation, while the second half corresponds to operation with anomalies.  The duration of the 327 anomalous periods for all machines constitute 4.2\% of the total duration of the dataset, with the maximum length of the anomaly window being 3161 points. 

\paragraph{SKAB} \emph{(Overrides defaults: transitions are anomalous; a 60s NAB tolerance window is applied.)} The SKAB dataset \cite{Katser2020} is a set of real-world sensor data collected from a water circulation testbed. The dataset comprises 35 experiments with 8 sensors, one is fault-free, and the others are structured as follows: 10 minutes of normal operation (400 points of each are in training), 1 minute transition to the anomaly, 5 minutes of steady anomaly, 1 minute recovery, and 3 minutes of normal post-event operation, and the data was measured at a frequency 1 Hz.  In this dataset we label the transition periods as anomalous. We evaluate change-point detection using a NAB-style tolerance window of 60 s.

\paragraph{TEP dataset} The TEP (Tennessee Eastman Process) dataset \cite{downs1993plant} is a synthetic benchmark widely used for fault detection and diagnosis tasks. This work utilizes the Reinartz version of TEP \cite{eTEP}, which includes 28 different fault types and 52 process variables. Data was sampled every 3 minutes. For each fault type, 100 simulations are conducted, partitioned into a training set (80 simulations) and a test set (20 simulations). As the process operates normally for the first 30 hours before transitioning to a faulty state for the remaining 70 hours, the proportion of anomalous data in each simulation is 70\%.

\begin{figure}
    \centering
    \includegraphics[width=0.9\linewidth]{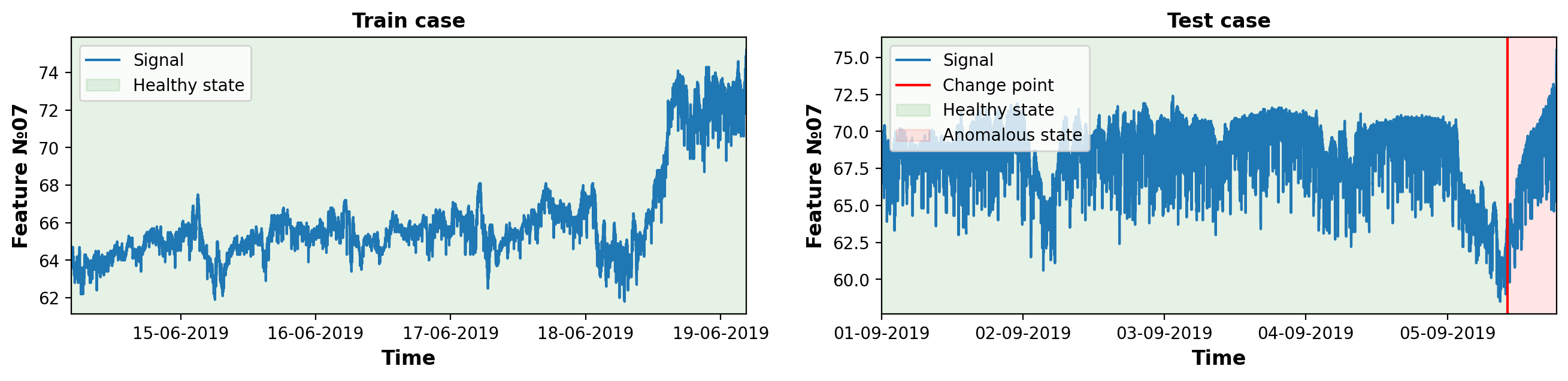}
    \caption{The plot of 7th feature from train and test sets in Proprietary dataset for Steam Turbine. }
    \label{fig:prophack_picture}
\end{figure}

\paragraph{Proprietary Dataset for Steam Turbine} \emph{(Overrides defaults: C1 trimmed to first 10 days due to exceptionally long anomaly.)}
The proprietary Turbine Dataset is a real-world dataset collected in 2019 at a frequency of 1 minute and consisting of two operational periods separated by a long shutdown: 20h and 8h with a gap $\approx$ 80 days and an anomaly rate $\approx$ 0.9\%. The periods span the entire degradation from onset to failure and are characterised by 70 features representing the physical parameters of the turbine. Throughout the dataset, multiple anomaly-related gaps in reading are observed, corresponding to operational downtimes caused by various failures. We define the train/test split: training uses only normal segments drawn from both operational periods; ; testing contains a short pre-failure normal context and all anomalous intervals (Fig.~\ref{fig:prophack_picture}).

\paragraph{NPPs turbogenerator proprietary dataset}

This dataset is a real-world dataset comprising logs for six multivariate cases (C1–C6) from nuclear turbogenerators, each containing a single continuous anomalous interval with a known change point. The total dataset duration is 445 days with 65 anomalous days; The anomaly rate varies from 7.6\% to 88.9\% with a median 11.6\%, and change points occur mostly late in time (median at 92.3\% of each sequence). Prior to resampling, cases contain 35,471–2,185,862 rows with heavy sparsity (minimum per-column missingness 50.1–81.2\%; 1–6 fully missing features). All series are resampled to 1-minute resolution, resulting in 18,780–168,617 samples per case (total 641,459; median 131,121). Train/test splits depend on the change point: training uses pre-change data, while testing includes a pre-change look-back and the anomalous interval; due to an exceptionally long anomaly, C1 is trimmed to its first 10 days before splitting. 

\begin{figure}
    \centering
    \includegraphics[width=1\linewidth]{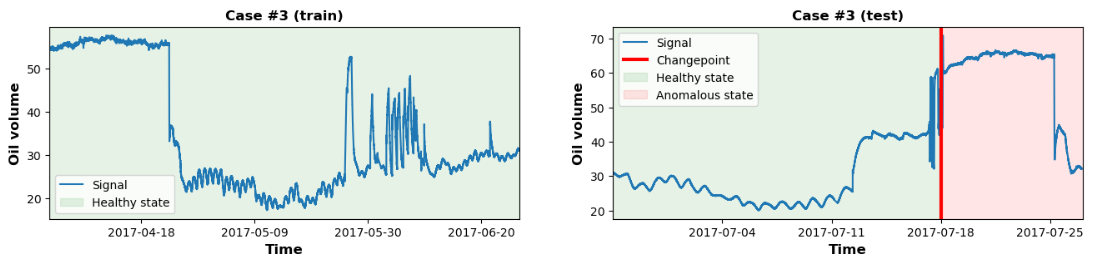}
    \caption{Oil volume for the third dataset divided into periods of healthy and anomalous states in the NPPs turbogenerator dataset}
    \label{fig:npp_oil}
\end{figure}

An example of data (for case 3) is shown in Fig.~\ref{fig:npp_oil}, extended data information are presented in Tab.~\ref{tab:npp_dataset}.

\begin{table}[t]
\centering
\caption{Characteristics of NPP datasets after preprocessing.}
\label{tab:npp_dataset}
\setlength{\tabcolsep}{3pt}
\begin{tabular}{lcccc}
\toprule
\makecell{Dataset \\ ID} &
\makecell{Features \\ (\#)} &
\makecell{Samples \\ ($\times10^3$)} &
\makecell{Duration \\ (days)} &
\makecell{Changepoint \\ (day)} \\
\midrule
\#1 & 221 & 39.0  & 27  & 4  \\
\#2 & 295 & 132.5 & 92  & 77 \\
\#3 & 225 & 168.6 & 117 & 108 \\
\#4 & 228 & 152.8 & 106 & 99 \\
\#5 & 228 & 129.8 & 90  & 83 \\
\#6 & 226 & 18.8  & 13  & 12 \\
\bottomrule
\end{tabular}
\end{table}

\paragraph{Preprocessing} We apply a unified preprocessing pipeline across all datasets: features are normalized using mean values and standard deviations from the training split, and then train dataset is sliced into fixed-length sliding windows. Evaluation additionally uses max-over-time anomaly score aggregation to assign score-based window labels.

Details of dataset-specific preprocessing are as follows: For \textbf{SWaT }, the data are segmented into runs based on timestamp gaps. For \textbf{SMD}, no dataset-specific cleaning is applied beyond optional subsetting by machine. For \textbf{WADI}, in addition  gap-aware segmentation, we discard non-informative and empty columns in the original data. However, the final number of columns corresponds to the number of sensors and actuators. For \textbf{TEP}, all 28 fault types are treated as a unified "anomaly" class. For \textbf{NPPs turbogenerator proprietary dataset} and \textbf{Proprietary Dataset for Steam Turbine} we imputed missing values with the previous known ones. In addition, we resampled \textbf{NPPs turbogenerator proprietary dataset} to 1-minute time interval between timestamps.

\section{Proposed approach}
\subsection{Event-level ablations}

\paragraph{Ablation-like preprocessing}

To simulate real-world problems, we implemented each preprocessing step as a separate module before the model input. Each module took an input tensor of shape [batch size, window length, feature length] and returned a processed tensor. We applied the following unified preprocessing pipelines. The values of the relative change parameters varied by dataset related coefficient:

\begin{itemize}

\item \textbf{Impact of noise on the model}  
We added Gaussian noise with zero mean and standard deviation equal to $n\%$ of the feature standard deviation. We applied the noise independently to each feature at every time step.



\item \textbf{Robustness to sensor failure and drift}  
We randomly selected 10\% of the feature channels and set their values to zero to model sensor failure. Then we applied a scale factor to the remaining channels to model gradual drift in sensor response.

\item \textbf{Impact of disabled sensors}  
We measured model performance as we turned off 0 – 10\% of the most important sensors. The number of disabled sensors scaled proportionally to the dataset size.

\item \textbf{Dynamic changes due to degradation}  
We applied two types of random drift to each feature channel. First, we multiplied each value at time step $t$ by $(1 + k\,t)$. Second, we multiplied by $(1 + k_{0}\,\log(k_{1}\,t))$ with randomly sampled coefficient. Each drift produced a gradual change over the full window.


\end{itemize}

\subsection{Evaluation Metric}
Let $(\mathrm{TP},\mathrm{FP},\mathrm{TN},\mathrm{FN})$ denote the counts of true positives, false positives, true negatives, and false negatives. During paper, we report only $F_1 = \frac{2\,\mathrm{TP}}{2\,\mathrm{TP} + \mathrm{FP} + \mathrm{FN}}$, however, the full metrics for each dataset are provided in the Supplementary Material.

\section{Models Under Investigation}
\label{sec:models}

 We evaluate fourteen recent anomaly‑detection architectures that together cover four design families: (i) \emph{sequence‑only encoders} that treat each multivariate series as a flat vector stream; (ii) \emph{graph‑augmented networks} that learn or assume sensor inter‑dependencies; (iii) \emph{density‑based generative models} that flag low‑likelihood regions; and (iv) a minimal \emph{linear autoencoder baseline}.  All models are trained on the same normal‑only splits and assessed with identical early‑stopping and hyper‑search budgets starting from original paper initializations. Their short description and references are presented in Tab. \ref{tab:model_summary}.


\begin{table*}[t]
    \centering
    \small
    \setlength{\tabcolsep}{4pt}
    \caption{Surveyed models and salient characteristics. Backbone codes: TR=Transformer, CNN=Convolutional network, RNN=Recurrent network, GNN=Graph neural net. Objective: Recon=reconstruction‑based, Forecast=forecasting‑based, Hybrid=both, Density=likelihood estimation.}
    \begin{tabularx}{\linewidth}{l c c X X}
        \toprule
        \textbf{Model} & \textbf{Backbone} & \textbf{Objective} & \textbf{Key Idea} & \textbf{Strengths / Limitations} \\
        \midrule
        AnomalyTransformer~\cite{AnomalyTransformer} & TR & Hybrid & Dual global vs.\ local (Gaussian) attention; discrepancy feeds loss & Captures mixed‑scale patterns; $\mathcal{O}(T^2)$ memory \\
        TimesNet~\cite{timesnet} & CNN & Recon & FFT‑guided periodic unfolding $\rightarrow$ Inception convs & Strong on seasonality; brittle under distribution shift \\
        MSCRED~\cite{mscred} & CNN+RNN & Recon & Signature matrices + ConvLSTM + attention & Multiscale correlation; dilutes sub‑second faults \\
        THOC~\cite{thoc} & RNN & Density & Dilated RNN + hierarchical one‑class clusters & Multi‑scale normality; long training time \\
        LSTM‑NDT~\cite{lstm_ndt} & RNN & Forecast & LSTM + dynamic thresholding & Deployed in spacecraft; may overfit gradual drift \\
        LSTM‑VAE~\cite{lstm_vae} & RNN & Recon & Variational AE with LSTM enc/dec & Uncertainty modelling; Gaussian assumption \\
        \midrule
        GDN~\cite{deng2021graph} & GNN+RNN & Forecast & Learnable graph + attention forecasting & Learns sensor topology; cost grows with sensors \\
        GBAD~\cite{gbad} & GNN & Recon & Graph‑structure learning layer + GCN encoder & Adaptive adjacency; lacks forecasting head \\
        GTA~\cite{chen2021learning} & GNN+TR & Forecast & Gumbel‑Softmax topology + multibranch attention & Reduces $\mathcal{O}(N^2)$ cost; GPU‑heavy \\
        MTAD‑GAT~\cite{zhao2020multivariate} & GNN+TR & Hybrid & Parallel temporal/variable GAT + VAE/MLP & Interpretable; balances two losses \\
        STGAT‑MAD~\cite{STGAT-MAD} & GNN+RNN & Recon & Multi‑scale 1D conv + dual graphs + BiLSTM AE & Local+global context; wide receptive field \\
        \midrule
        GANF~\cite{ganf} & Flows & Density & DAG factorisation with normalising flows & Causal insight; scaling in high $d$; costly \\
        USAD~\cite{audibert2020usad} & AE+GAN & Recon & Two‑decoder adversarial autoencoder & Simple; adversarial instability \\
        MLPREC~\cite{MLPREC} & Linear & Recon & 2‑layer linear autoencoder baseline & Fast; low expressivity \\
        \bottomrule
    \end{tabularx}    
    \label{tab:model_summary}
\end{table*}


\section{Experiments and Discussions}
We conduct a systematic evaluation of recent anomaly detection models under both standard and stress conditions, aiming to assess realistic deployment challenges. Our benchmark covers several open and proprietary datasets, each capturing different degradation behaviors and operational regimes. For details on the reproduction of prior results on open datasets and related checkpoints metrics used in downstream analyses, see the \textit{Supplementary} section.

\begin{table*}[t]
\centering
\small
\setlength{\tabcolsep}{5pt}
\caption{F1-scores across datasets (higher is better). Best per dataset in \textbf{bold}. Unless stated otherwise, differences $\le$0.01--0.02 F$_1$ points are not statistically significant across three seeds.}
\begin{tabular}{lccccccc}
\toprule
\textbf{Model} & \textbf{SKAB} & \textbf{TEP} & \textbf{WADI} & \textbf{SWaT } & \textbf{SMD} & \textbf{Turbogenerator} & \textbf{Steam Turbine}\\
\midrule
AnomalyTransformer  & 0.84  & 0.849 & 0.19  & 0.76  & 0.64  & 0.935 & 0.407 \\
DAGMM               & 0.75  & 0.83  & 0.23  & 0.75  & 0.22  & 0.877 & 0.382 \\
GANF                & \textbf{0.87} & 0.87  & 0.47  & 0.80  & 0.55  & 0.883 & 0.39  \\
GBAD                & 0.81  & 0.892 & 0.425 & 0.804 & 0.709 & 0.920 & 0.384 \\
GDN                 & 0.77  & 0.825 & 0.539 & \textbf{0.815} & 0.41  & 0.776 & 0.385 \\
GTA                 & 0.807 & 0.888 & 0.468 & 0.767 & 0.644 & 0.903 & 0.411 \\
LSTM\textendash VAE & 0.82  & \textbf{0.91}  & 0.47  & 0.767 & 0.63  & 0.870 & 0.50  \\
MLPREC              & 0.82  & 0.890 & 0.418 & 0.775 & 0.701 & 0.906 & 0.307 \\
MSCRED              & 0.83  & 0.84  & 0.39  & 0.761 & 0.54  & \textbf{0.975} & 0.40  \\
MTAD\textendash GAT & 0.73  & 0.85  & 0.43  & 0.76  & 0.61  & 0.894 & 0.411 \\
STGAT\textendash MAD$^\dagger$ & 0.81  & 0.896 & 0.540 & 0.76  & 0.637 & 0.904 & \textbf{0.543} \\
THOC                & \textbf{0.87} & 0.90  & 0.56  & 0.72  & 0.59  & 0.895 & 0.441 \\
TimesNet            & 0.85  & 0.85  & \textbf{0.719} & 0.76    & \textbf{0.72}  & 0.893 & 0.19  \\
USAD                & 0.80  & 0.834 & 0.339 & 0.763 & 0.54  & 0.869 & 0.343 \\
\bottomrule
\end{tabular}
\begin{flushleft}
\end{flushleft}
\label{tab:all_f1_summary}
\end{table*}

\paragraph{Standard benchmarking protocols}
On open benchmarks, leadership is dataset-specific and often tightly clustered, with WADI the main exception showing a single dominant family. Scores are clean, event-level F1 with thresholds selected on validation and fixed before testing. On proprietary telemetry, outcomes polarize: Turbogenerator shows a ceiling effect with many models performing similarly well, whereas Steam Turbine is distinctly hard, with uniformly lower scores and only a graph-based approach at the top.

The experiment results suggest an actionable mapping: (i) when periodicity is strong and phases are stable (WADI, SMD), spectral models (e.g., TimesNet) lead by a large margin; (ii) on clean, stationary plants (SKAB, TEP), density/flow or low-rank/reconstruction models (e.g., THOC, GANF, LSTM–VAE) are competitive within <0.02 F1 of graph baselines; (iii) under industrial missingness and long events (Steam Turbine), explicit topology (e.g., STGAT) is currently the only family at the top, while generic predictors and spectral/density models underperform.

\paragraph{Sensors impact}

\begin{figure}[t]
  \centering
  \begin{subfigure}[t]{0.49\linewidth}
    \centering
    \includegraphics[width=\linewidth]{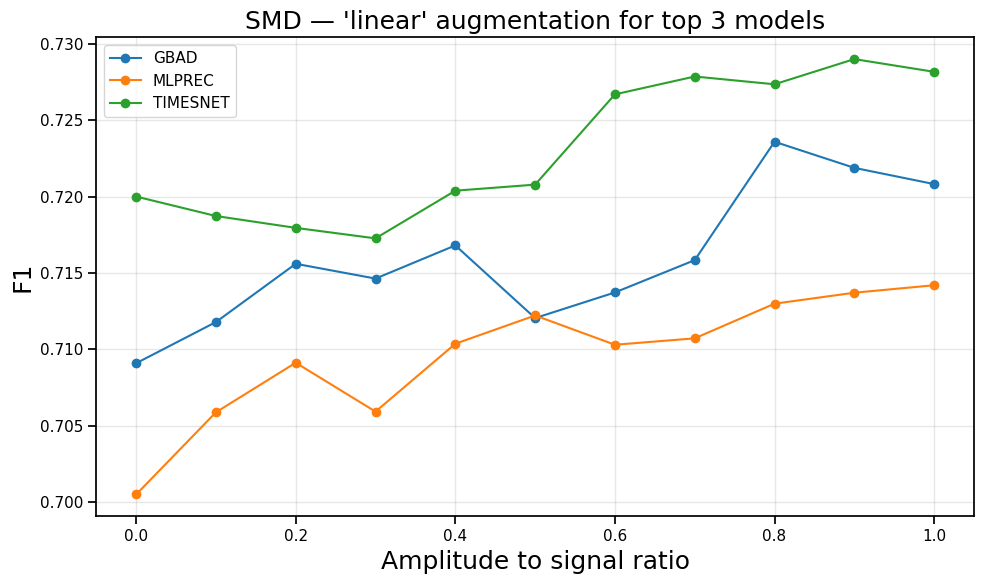}
    \caption{SMD — \texttt{linear}}
    \label{fig:smd-linear}
  \end{subfigure}\hfill
  \begin{subfigure}[t]{0.49\linewidth}
    \centering
    \includegraphics[width=\linewidth]{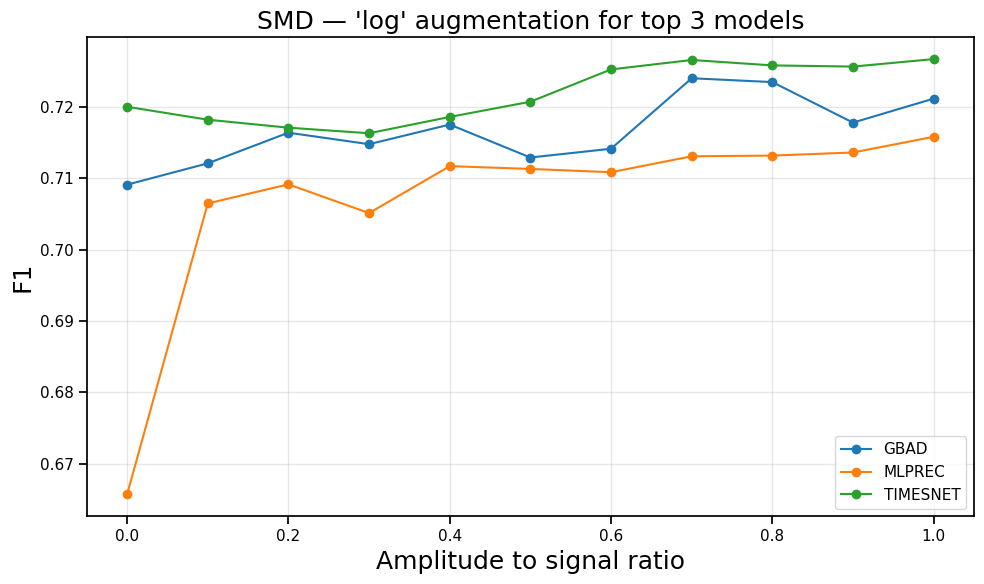}
    \caption{SMD — \texttt{log}}
    \label{fig:smd-log}
  \end{subfigure}

  \vspace{0.5em}

  \begin{subfigure}[t]{0.49\linewidth}
    \centering
    \includegraphics[width=\linewidth]{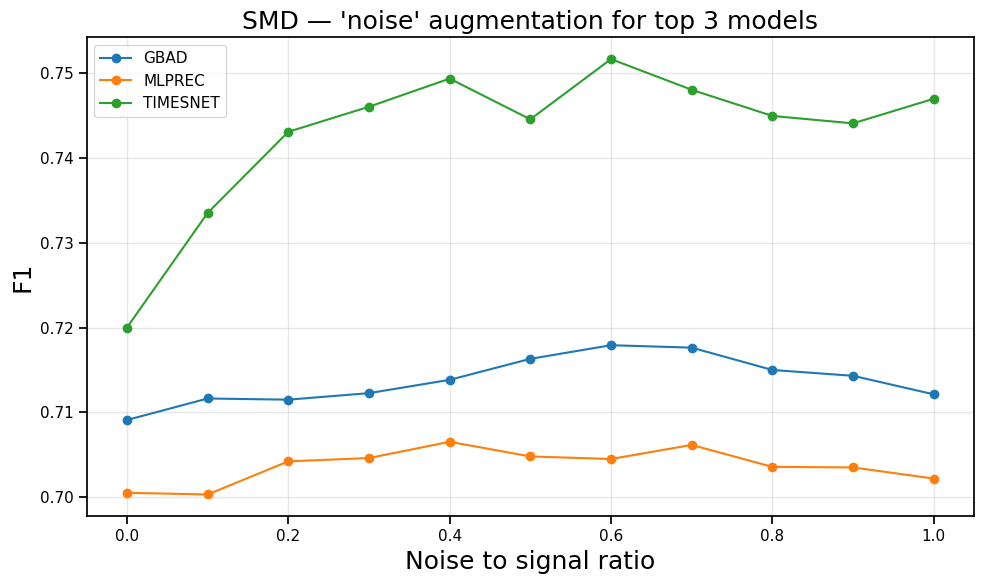}
    \caption{SMD — \texttt{noise}}
    \label{fig:smd-noise}
  \end{subfigure}\hfill
  \begin{subfigure}[t]{0.49\linewidth}
    \centering
    \includegraphics[width=\linewidth]{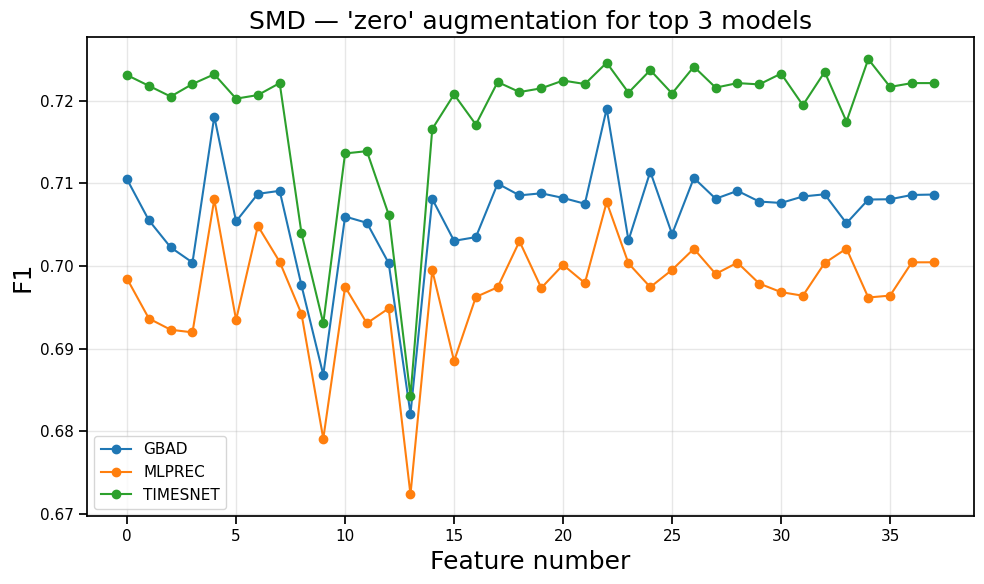}
    \caption{SMD — \texttt{zero-channel}}
    \label{fig:smd-zero}
  \end{subfigure}

  \caption{Top-3 models on SMD under four stress augmentations simulated industrial events.}
  \label{fig:smd-aug-grid}
\end{figure}

\begin{figure}[t]
  \centering
  \begin{subfigure}[t]{0.49\linewidth}
    \centering
    \includegraphics[width=\linewidth]{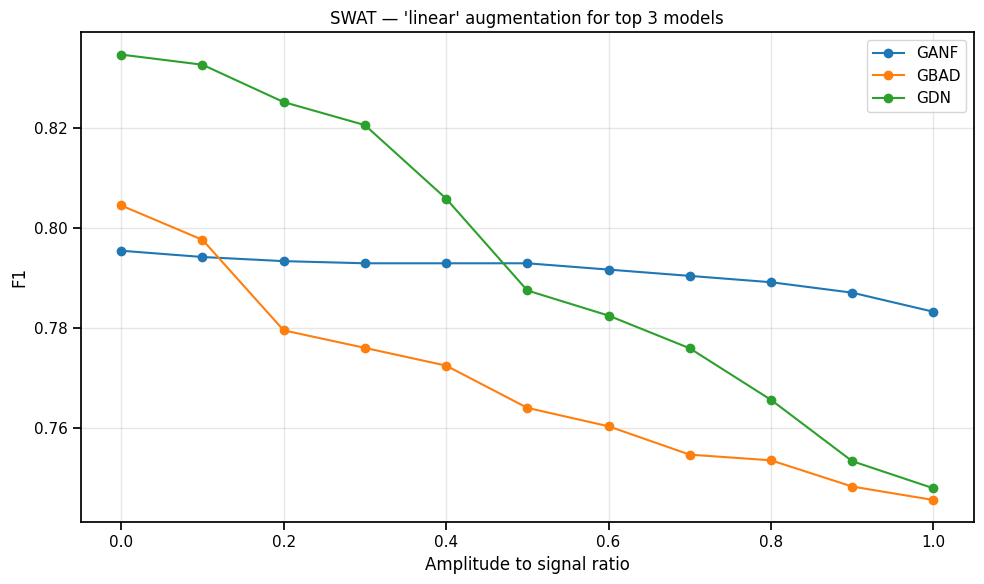}
    \caption{SWaT  — \texttt{linear}}
    \label{fig:swat-linear}
  \end{subfigure}\hfill
  \begin{subfigure}[t]{0.49\linewidth}
    \centering
    \includegraphics[width=\linewidth]{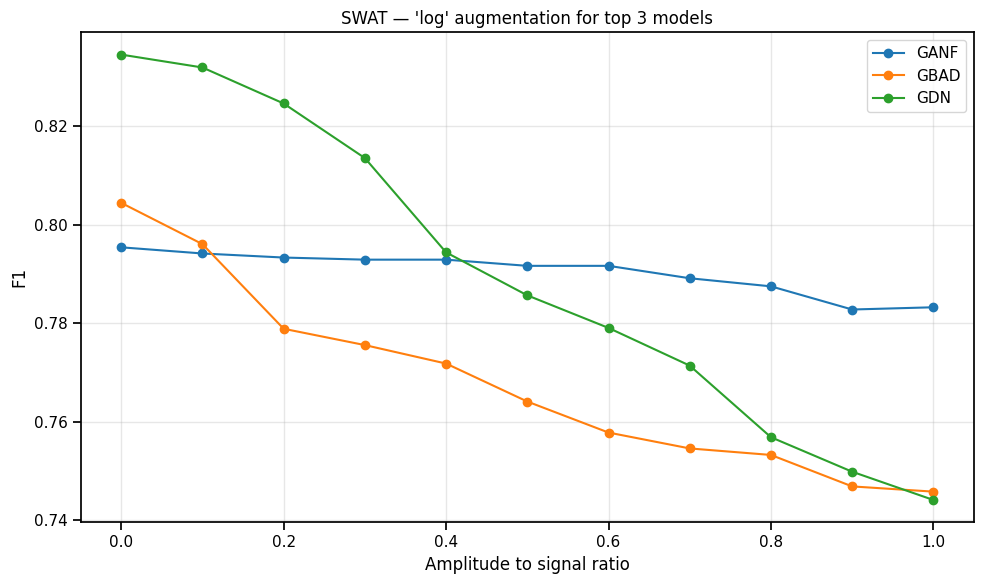}
    \caption{SWaT  — \texttt{log}}
    \label{fig:swat-log}
  \end{subfigure}

  \vspace{0.5em}

  \begin{subfigure}[t]{0.49\linewidth}
    \centering
    \includegraphics[width=\linewidth]{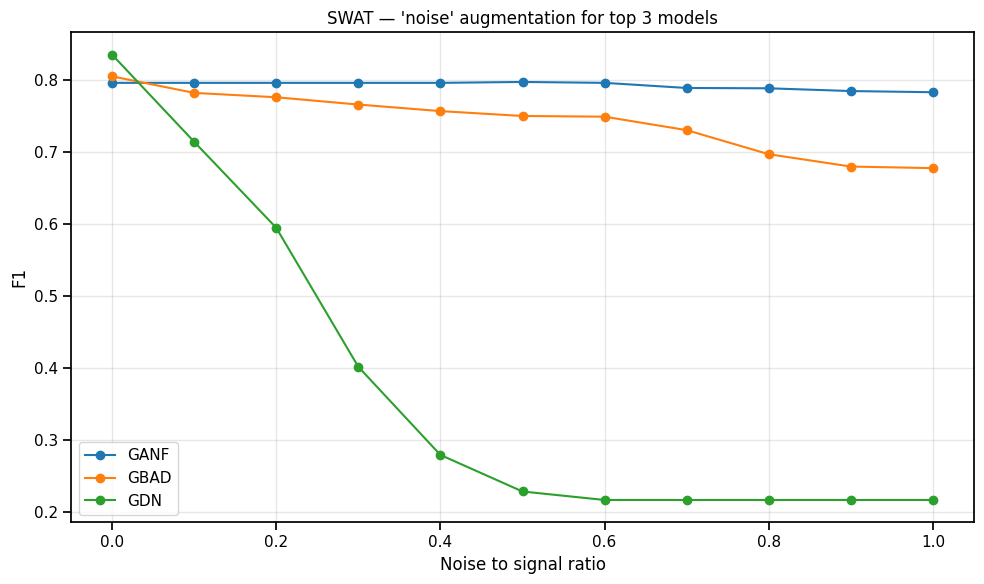}
    \caption{SWaT  — \texttt{noise}}
    \label{fig:swat-noise}
  \end{subfigure}\hfill
  \begin{subfigure}[t]{0.49\linewidth}
    \centering
    \includegraphics[width=\linewidth]{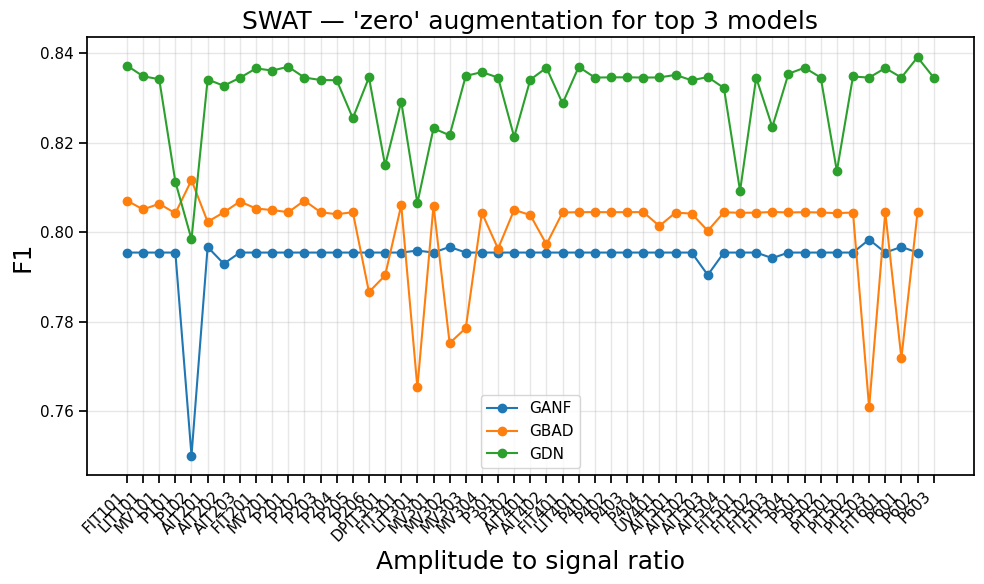}
    \caption{SWaT  — \texttt{zero-channel}}
    \label{fig:swat-zero}
  \end{subfigure}

  \caption{Top-3 models on SWaT under four stress augmentations simulated industrial events.}
  \label{fig:swat-aug-grid}
\end{figure}

On SMD, trend-type augmentations (linear, log) have only mild effects: GBAD and TimesNet trace nearly flat, slightly rising curves, while the MLP reconstruction baseline jumps early and then plateaus, indicating low sensitivity to slow drifts. Under additive noise, TimesNet improves up to a mid-range peak before tapering; GBAD moves only a few basis points; and MLPREC shows a small regularization-like gain—overall a stable trio. Zero-channel probing reveals sharp per-channel valleys clustered around the same feature index for all three models (deepest for MLPREC, visible for GBAD, and a narrow dip for TimesNet), pointing to a single toxic or over-dominant sensor that perturbs ranking locally but not globally.

On SWaT, trends drive a clear separation: GDN degrades monotonically and strongly as trend strength increases, GBAD declines steadily, and GANF remains almost flat with the smallest loss. Under additive noise, the contrast is starker: GANF stays near baseline, GBAD decays gradually with noise level, and GDN collapses after modest noise—consistent with variable-attention pipelines being sensitive to unmodeled perturbations. Zero-channel curves are largely flat for GANF and GDN, with occasional dips for GBAD around a few channels, suggesting limited ranking changes without explicit sensor vetting.

\paragraph{Takeaways.}
(i) For stable plants with moderate drift and noise (SWaT -like regimes), flow-based density models are the most noise-tolerant; graph autoencoders are acceptable but erode with stress; and variable-attention models are sensitive. (ii) For SMD-like workloads, the top models transfer well under slow drifts; spectral CNNs can even benefit from modest noise, likely via frequency pooling acting as a denoiser. (iii) Sensor-level probing is actionable: a few channels dominate failure modes and can flip local rankings—basic sensor vetting or zeroing should precede model selection. Overall, there is no universal winner; match inductive bias to the stress profile (flows \(\leftrightarrow\) stable/noisy, graph structure \(\leftrightarrow\) missingness/long events, attention/hybrid \(\leftrightarrow\) dynamic breaks with careful windowing) and sanity-check channels before deployment.

\paragraph{Sensor drop + shift} Across stress-composition tests we observe dominated degradation: for a given model, either drift-like or missingness-like perturbations govern performance, and the combined stress yields an F1 decrease comparable to the larger single-stressor drop at the same severity rather than a sum of both. In other words, the overall fall is on the order of the per-stressor experiments, not super-additive, under our fixed-severity setting.

For example, on SMD with a fixed linear additive perturbation of amplitude 0.4 combined with five 10\% random sensor dropouts (averaged across masks), with thresholds fixed on validation and event-level scoring, we obtain: TimesNet 0.711 (0.008) and MLPREC 0.707 (0.011). Relative to the clean baselines from Table~\ref{tab:all_f1_summary} (TimesNet 0.720; MLPREC 0.701), TimesNet decreases by 0.009 points (about 1.25\%), while MLPREC increases by 0.0058 points (about 0.82\%). Both models remain within roughly 1–1.5\% of their clean scores; the mean gap between them under stress is small compared to across-mask variability, so we do not claim statistical significance. These results concretely illustrate that the net effect of composing drift with sensor dropout is dominated by the model’s primary sensitivity rather than additive collapse.

\begin{figure}[t]
  \centering
    \centering
    \includegraphics[width=\linewidth]{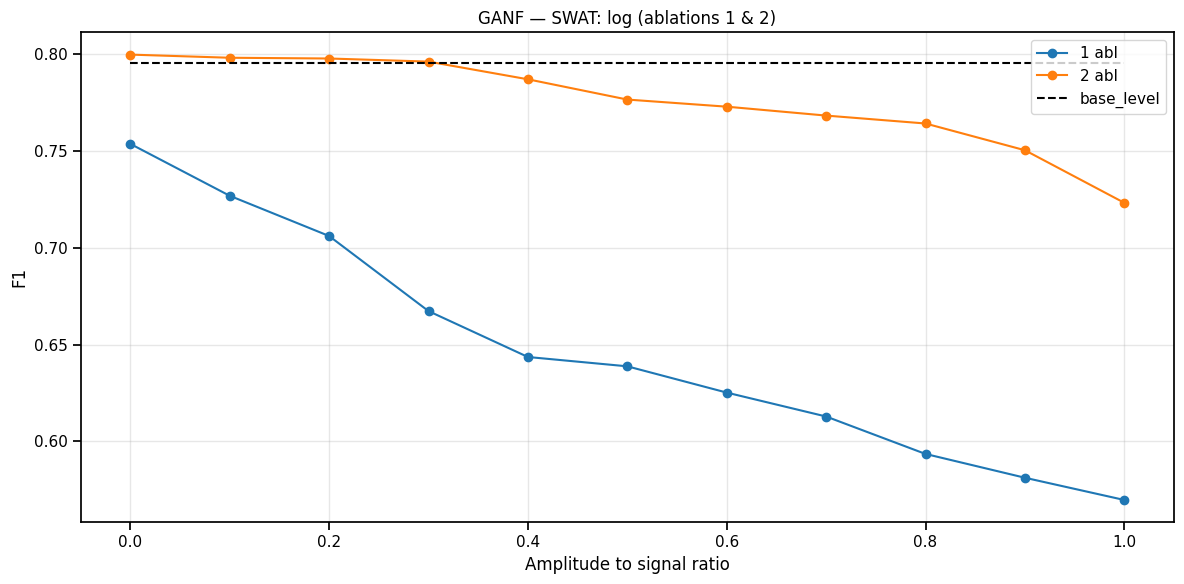}
  \caption{GANF model under replacing normalizing flows (1) with a Gaussian density estimator (GDE) and changing the learned DAG to fixed graph (2).}
  \label{fig:ganf-swat-log}
\end{figure}

\paragraph{Design usage, studying speed optimization of GANF on the SWaT  dataset} 
We make an example of using proposed benchmarking approach to study the relative contribution of architectural blocks while pursuing GANF speedups (Fig. ~\ref{fig:ganf-swat-log}). 
For the original model on SWaT , performance is nearly invariant to a monotone log transform (total drop \(\sim 0.01\text{--}0.02\)). 
We then probe two speed\textendash oriented ablations: 
\emph{(i)} replace normalizing flows with a Gaussian density estimator (GDE) and 
\emph{(ii)} remove the learned DAG (use a fixed graph). 
The fixed\textendash graph variant is slightly higher on the clean point (\(+0.5\%\text{--}1\%\)), which under a clean\textendash set benchmark could be reported as new state-of-the-art; 
\emph{however}, its stress curve is much steeper, falling to \(F_1 \approx 0.72\) (a loss of \(0.07\text{--}0.08\) vs.\ \(\approx 0.01\) for the original), 
i.e., far more sensitive to connectivity shifts. 
The GDE variant shows a qualitatively different—and notably undesirable—profile: accuracy declines rapidly with stress 
(from \(F_1 \approx 0.75\) at \(0\) to \(\approx 0.57\) at \(1.0\); already \(\approx 0.70\) at \(0.2\) vs.\ \(\approx 0.79\) baseline). 
Learned flows and a learned graph are the components that buy robustness to monotone sensor transforms and naïve speedups risk large losses under realistic drift. 

\paragraph{Findings}

\begin{itemize}
  \item \textbf{Graph-structured / graph-attention (STGAT, MTAD-GAT, GBAD).}
  On SWaT  with additive noise, graph autoencoders degrade more than hybrid graph models:
  GBAD \( \approx 0.804 \rightarrow 0.677 \) (\(-16\%\), \(Recall\!\approx\!0.84\), see Supplementary),
  STGAT \( \approx 0.759 \rightarrow 0.680 \) (\(-10\%\)),
  while MTAD-GAT stays nearly flat (\(0.762 \rightarrow 0.756\), \(-0.8\%\)).
  Under linear/log drift the drop is moderate—about \(-7\%\) for GBAD and \(-5\%\ldots-6\%\) for STGAT—consistent with the advantage of explicit topology for long episodes, but a sensitivity to noise.

  \item \textbf{Density / flow (GANF).}
  On clean, stationary plants they retain performance even under noise (NPP/TEP: Recall $(\ge 0.99\,)$, changes within \(\pm1\%\), see Supplementary),
  but log-drift yields catastrophic cases: on SKAB and NPP scores collapse toward \(0.0\) at small drift;
  on SWaT  the decline is mild (\(0.795 \rightarrow 0.783\), \(-1.5\%\)), and on TEP about \(-4\%\).
  This underscores how strongly flows rely on stationarity/factorization assumptions.

  \item \textbf{Spectral / seasonal CNNs (TimesNet-class).}
  On the Turbogenerator dataset performance is low and unstable: noise reduces scores by roughly \(10\%\ldots15\%\) (\(\approx 0.20 \rightarrow 0.17\)),
  and zero-channel probing shows only local fluctuations without consistent gains.
  Best when seasonality is well defined; degrade under noise/drift.

  \item \textbf{Reconstruction AEs (MSCRED, USAD, LSTM-VAE).}
  On SWaT /TEP noise curves are nearly flat (e.g., USAD@SWaT : \(0.763 \rightarrow 0.755\), \(-1\%\);
  MSCRED@TEP: changes \(<0.2\%\)),
  but transferability to ``rough'' datasets is weaker (MSCRED@WADI \(\approx 0.14\)).
  Sensor-level probing can also confirm the role of toxic channels: on the Turbogenerator dataset, disabling it lifts GBAD scores from \(\approx 0.38 \rightarrow 0.58\) (\(+54\%\)).

  \item \textbf{Predictive / hybrid dynamics (Anomaly Transformer, MTAD-GAT).}
  Strong on dynamic faults but noise-sensitive: on the Turbogenerator dataset, Anomaly Transformer drops \(\approx 0.407 \rightarrow 0.312\) (\(-23\%\)).
  Zero-channel probing shows both improvements (up to \(+6\%\)) on informative channels and regressions (down to \(-18\%\)) on dominant/toxic ones—evidence of channel over-reliance and the value of sensor vetting.

  \item \textbf{Model-by-dataset counterpoints (no universal winner).}
  THOC is almost unchanged on TEP (\(\approx 0.90\), \(Recall\!\approx\!1\), see Supplementary) but falls by \(45\%\ldots70\%\) on SWaT /WADI under noise—the same architecture behaves very differently across regimes.
  On SWaT  noise, GBAD degrades more than MTAD-GAT (\(-16\%\) vs. \(-0.8\%\)), yet under long episodes/drift their gap narrows (\(-5\%\ldots-7\%\)).
  \emph{Sensor-level reshuffle:} on the Turbogenerator dataset, zero-channel boosts GBAD by up to \(+54\%\) (\(\approx 0.38 \rightarrow 0.58\)), while Anomaly Transformer spans \(-18\%\ldots+6\%\) across channels; clean sets rankings are unstable once root-cause probing is applied.
\end{itemize}

\section{Conclusion}
We introduced a deployment-first, \emph{event-level} protocol for multivariate CPS anomaly detection that enforces \emph{zero test-time calibration} and adds an offline-calibrated stress suite plus sensor-level probing. Across \textbf{14} models on \textbf{7} datasets, there is no universal winner and rankings flip under realistic perturbations; e.g., on SWaT with additive noise a graph autoencoder loses $\sim\!16\%$ F$_1$, while a hybrid graph–attention stays nearly flat. These results turn leaderboards into \emph{design rules}: prefer graph structure for missingness/long events, spectral CNNs for stable periodicity, density/flow or reconstruction for clean stationary plants, and predictive/hybrid dynamics when faults break temporal dependencies (mind window sensitivity). Replacing learned flows/graphs with faster surrogates erodes robustness under drift.

\noindent\textbf{Limitations \& next steps.} Stress severity is normalized by validation statistics and event definitions are fixed. Extending to benchmark embedded data-specific stress testing is promising.

\section{Acknowledgements}

The study was supported by the Ministry of Economic Development of the Russian Federation (agreement No. 139-10-2025-034 dd. 19.06.2025, IGK
000000C313925P4D0002)

\bibliography{bibliography}
\end{document}